\documentclass[letterpaper, 10 pt, journal, twoside]{IEEEtran}
\usepackage{graphicx}
\usepackage{cite}
\usepackage{url}
\usepackage{colortbl}
\usepackage{multirow}
\usepackage{amssymb}
\usepackage{mathtools}
\usepackage{bbm}
\usepackage{xcolor}
\usepackage{booktabs}

\usepackage{hyperref}
\hypersetup{
    colorlinks=true,
    linkcolor=red,
    filecolor=magenta,      
    urlcolor=blue,
}

\definecolor{limegreen}{rgb}{0.2, 0.8, 0.2}
\definecolor{forestgreen}{rgb}{0.13, 0.55, 0.13}
\definecolor{greenhtml}{rgb}{0.0, 0.5, 0.0}
\definecolor{black}{rgb}{0.0, 0.0, 0.0}

\usepackage[font=small,labelfont=bf,tableposition=top]{caption}

\usepackage{blindtext}

\definecolor{tableone}{RGB}{255, 191, 191}
\definecolor{tabletwo}{RGB}{255, 241, 1}
\definecolor{limegreen}{rgb}{0.2, 0.8, 0.2}
\definecolor{forestgreen}{rgb}{0.13, 0.55, 0.13}
\definecolor{greenhtml}{rgb}{0.0, 0.5, 0.0} 

\let\oldtwocolumn\twocolumn
\renewcommand\twocolumn[1][]{%
    \oldtwocolumn[{#1}{
    \begin{center}
           \includegraphics[width=17cm]{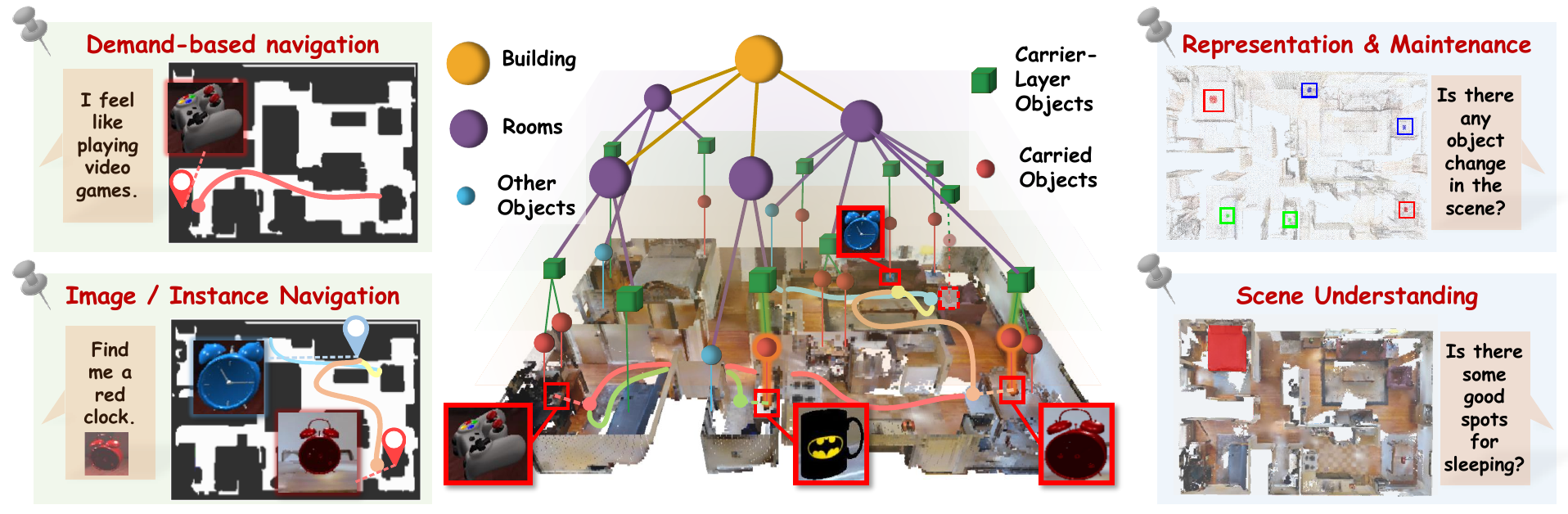}
           \captionof{figure}{\textcolor{black}{OpenIN supports multi-modal (text and image) and multi-type (demand, semantic, and instance-level) object navigation. Additionally, it enables maintaining 
           scene representations through a dynamic carrier-relationship scene graph, along with scene understanding.}}
           \label{first}
        \end{center}
    }]
}

\begin{document}
\pagestyle{plain}
\title
{
{OpenIN: Open-Vocabulary Instance-Oriented Navigation in Dynamic Domestic Environments}

\author{
	{Yujie Tang$^1$, Meiling Wang$^1$, Yinan Deng$^{1}$, Zibo Zheng$^{2}$, Jingchuan Deng$^{1}$, Yufeng Yue$^{1*}$}}

\thanks{
This work is  supported by the National Natural Science Foundation of China under Grant No. NSFC 62233002, 62003039.
(Corresponding Author: Yufeng Yue, yueyufeng@bit.edu.cn)}

\thanks{$^1$Yujie Tang, Meiling Wang, Yinan Deng, Jingchuan Deng, Yufeng Yue  are with
School of Automation, Beijing Institute of Technology, Beijing, 100081, China.
}

\thanks{$^2$Zibo Zheng  are with School of Mechanical Engineering, University of Nottingham Ningbo China, Ningbo, 315100, China.}
}

\maketitle

\begin{abstract}
In daily domestic settings, frequently used objects like \textit{cups} often have unfixed positions and multiple instances within the same category, and their carriers frequently change as well. As a result, it becomes challenging for a robot to efficiently navigate to a specific instance. To tackle this challenge, the robot must capture and update scene changes and plans continuously. However, current object navigation approaches primarily focus on the semantic level and lack the ability to dynamically update scene representation.  In contrast, this paper captures the relationships between frequently used objects and their static carriers. It constructs an open-vocabulary Carrier-Relationship Scene Graph (CRSG) and updates the carrying status during robot navigation to reflect the dynamic changes of the scene. Based on the CRSG, we further propose an instance navigation strategy that models the navigation process as a Markov Decision Process. At each step, decisions are informed by the Large Language Model's commonsense knowledge and visual-language feature similarity.
We designed a series of long-sequence navigation tasks for frequently used everyday items in the Habitat simulator. The results demonstrate that by updating the CRSG, the robot can efficiently navigate to moved targets. Additionally, we deployed our algorithm on a real robot and validated its practical effectiveness. The project page can be found here: \url{https://OpenIN-nav.github.io}.
\end{abstract}

\section{Introduction}
\label{introduction}
As one of the fundamental tasks in embodied AI, object navigation \cite{zhang20233d} has garnered widespread attention from researchers.  Imagine a daily environment in which a robot is tasked with navigating efficiently to any object, whether it is static furniture or a frequently used item with changing positions, such as \textit{a black cup}. \textcolor{black}{This poses several requirements for the navigation algorithm: 1. Open-vocabulary\cite{peng2023openscene} recognition and instruction, 2. Accurate instance differentiation, 3. Memorizing and updating object states, and 4. Effective navigation strategies. However, there is still a significant gap in achieving this goal.}

Many existing object navigation methods\cite{chang2020semantic, ye2021auxiliary,chaplot2020object,luo2022stubborn,ramakrishnan2022poni,zhang20233d,rajvanshi2024saynav} are limited to  closed-set navigation, leading to poor performance with unknown objects. With the advancement of visual language models (VLM) \cite{zhang2024vision} and large language models (LLM)\cite{chang2024survey},  open-vocabulary detection is achievable. However,  most object navigation methods\cite{chen2023not,zhou2023esc,yokoyama2024vlfm} support only one predefined instruction type, limiting flexibility. In contrast, humans use diverse instructions, such as demands (e.g., ``\textit{I'm hungry}") or instance descriptions. To meet this requirement, the proposed method supports multiple instruction types, including demands, semantics, and instance descriptions (e.g., color or carried-by relationships). 
In addition, instance-level navigation requires precise target discrimination, but most methods lack dedicated target identification modules, resulting in lower accuracy. To address this, we designed a text-image multimodal input, combining text feature similarity, RGB feature similarity, and image-level similarity from a visual-language model to enhance the accuracy of instance-level object navigation.

Moreover, in daily scenarios, the location of frequently used instance objects (e.g., \textit{a blue cup}) is dynamic (e.g., from \textit{ the living room} to \textit{the kitchen}), and their carriers can change (e.g., \textit{on a coffee table} or \textit{dining table}), making efficient navigation to specific instances challenging. This requires scene memory and dynamic updates, which many existing methods struggle to handle effectively\cite{chang2020semantic, ye2021auxiliary,chaplot2020object,luo2022stubborn,ramakrishnan2022poni,zhang20233d, dorbala2023can, chen2023not, zhou2023esc,yokoyama2024vlfm,rajvanshi2024saynav}. We capture the carried-by relationships between commonly used instance objects and their carriers, constructing a dynamic Carrier-Relationship Scene Graph (CRSG) to memorize carrying and carried-by relationships and update the states of instance objects, enabling promising lifelong navigation. 

Finally, the positional variability of commonly used instances requires an efficient navigation strategy. Based on the CRSG, we designed an instance-oriented navigation strategy that models the object search process as a Markov Decision Process (MDP) \cite{garcia2013markov}. At each navigation step, the robot navigates toward a candidate target or an unexplored carrier object. Specifically, we use visual-language feature similarity to identify candidate targets, and consider factors such as navigation cost and feature similarity to rank the navigation order of them. If no candidate target is available, we rank the exploration order of carrier objects based on commonsense knowledge provided by the large language model (e.g., \textit{a cup is more likely to be placed on the table}), enabling efficient instance navigation.

In summary, our contributions are as follows:
\begin{itemize}
\item We present an open-vocabulary, instance-oriented navigation system that supports multi-modal and multi-type object navigation instructions, enabling effective navigation to \textcolor{black}{everyday instances with variable positions.}
\item We present an adaptable carrier relationship scene graph (CRSG) that primarily describes the dynamic carried-by relationships between objects.
\item We design a navigation strategy based on the CRSG, utilizing visual-language features and commonsense knowledge from the LLM to inform decision-making.
\item Extensive qualitative and quantitative experiments demonstrate that updating the CRSG contributes to efficient navigation in tasks involving long sequences of moved instances. Additionally, we deployed the algorithm on a real robot, confirming its practicality.
\end{itemize}

\section{Related Work} \label{related work}

\subsection{Open Vocabulary Mapping}
With the advent of vision-language models like CLIP\cite{radford2021learning} and its variants, scene reconstruction  has moved beyond the limitations of fixed classes\cite{deng2022s}, and expanded to open-vocabulary \cite{ peng2023openscene, gu2024conceptgraphs}. Clip-fields\cite{shafiullah2022clip} integrates CLIP and SBERT features into the neural implicit map, enabling open-vocabulary map queries and navigation for a robot. Vlmap\cite{huang2023visual} projects CLIP features top-down onto a 2D grid to enable zero-shot language navigation. Conceptgraph\cite{gu2024conceptgraphs} and Hovsg\cite{werby23hovsg} constructed  instance-level point cloud maps with CLIP features embedded and scene graph representing certain relationships between objects, which facilitates more detailed and precise object retrieval.

These maps provide crucial support for applications such as open-vocabulary object queries, scene understanding, and robot navigation. However, they generally lack dynamic update capabilities. We construct a dynamic carrier-relationship scene graph (CRSG) that describes the dynamic carrier-carried relationships between objects, and continuously update the CRSG during navigation. This enables more efficient navigation to everyday instances.

\subsection{Object Navigation}
Object navigation\cite{chang2020semantic, ye2021auxiliary,chaplot2020object,luo2022stubborn,ramakrishnan2022poni,zhang20233d, rajvanshi2024saynav, dorbala2023can, chen2023not,zhou2023esc,yokoyama2024vlfm, kuang2024openfmnav}, as one of the key tasks in the field of embodied AI, primarily involves navigating to a specified semantic object or instance within a scene. \cite{chang2020semantic, ye2021auxiliary,chaplot2020object,luo2022stubborn,ramakrishnan2022poni,zhang20233d,rajvanshi2024saynav}  mainly perform object navigation within the closed-set classes.
\cite{gu2024conceptgraphs, huang2024ivlmap, werby23hovsg} constructed an open-vocabulary instance map of the scene based on VLMs, enabling open-set instance navigation. \cite{dorbala2023can, chen2023not, zhou2023esc,yokoyama2024vlfm}  perform open-vocabulary object navigation using the frontier exploration method. However, they\cite{gu2024conceptgraphs, huang2024ivlmap, werby23hovsg,dorbala2023can, chen2023not, zhou2023esc,yokoyama2024vlfm} are unable to capture and update the dynamics of instances in everyday environments and typically lack an instance discrimination module, which makes efficient navigation to dynamic instances challenging. In contrast, we build a dynamic CRSG that captures instance changes. Moreover, by considering multiple factors, such as visual-language features and image similarity, we achieve more accurate instance identification and navigation. The approach most similar to ours is GOAT\cite{chang2023goat}, which also implements memory capabilities for the latest scene and supports multi-type and multi-modal navigation command inputs. However, for navigating to a displaced everyday object, we designed a  navigation strategy based on CRSG, while GOAT selects the closest unexplored region to navigate to the object.

\begin{figure*}[ht]
\centering
{\includegraphics[width =17cm]{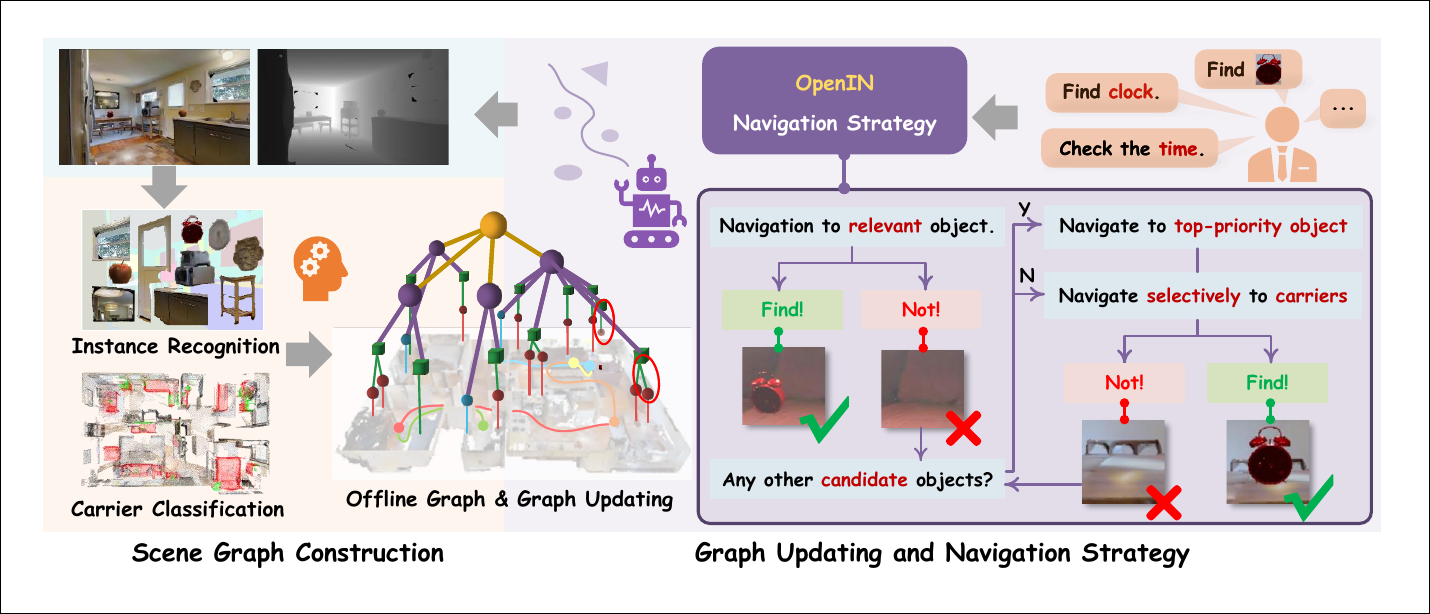}}
\caption{
\textcolor{black}{The OpenIN framework consists of two main modules.} The Scene Graph Construction module focuses on constructing the scene graph that describes the carrier-carried relationships. The Graph Updating and Navigation Strategy module is responsible for executing cognitive navigation based on user instructions, following the proposed navigation strategy, while updating the scene graph in the process.
}
\setlength{\belowcaptionskip}{-0.2cm}
\label{system-framework}
\end{figure*} 

\textcolor{black}{\section{Method} \label{Systematic Framework}}

\subsection{Problem Definition}
In a daily environment, when given a navigation command, \textcolor{black}{an intuitive answer is that} the robot queries the offline map to determine the endpoint and navigates there. If the target is a daily item (e.g., \textit{a cup}) that is being carried, the robot evaluates whether the item remains in its original location based on the current observations. If not, the robot initiates a strategic exploration process. We define this challenge as a \textbf{displaced instance exploration and navigation task} within an everyday setting. An overview of the framework is provided in Fig.
\ref{system-framework}.



\subsection{Carrier-Relationship Scene Graph (CRSG)}
We first construct an open-vocabulary instance map $\mathcal{M}$ using the pre-collected RGB-D data of the scene. Unlike Conceptgraph\cite{gu2024conceptgraphs}, where each instance object $\textit{\textbf{O}}_i \in \textit{\textbf{O}}$ (\textit{\textbf{O}} is the set of all objects) is represented by a CLIP feature $\textit{\textbf{V\_F}}_i$, we enhance each instance by adding a caption description list $\textit{\textbf{cap}}_i$, generated using the Tokenize Anything model \cite{pan2023tokenize}, as well as a text feature $\textit{\textbf{T\_F}}_i$ encoded using the SBERT model \cite{reimers2019sentence}. A Carrier-Relationship Scene Graph (CRSG) $S\_G$ is then constructed below.

\textcolor{black}{\textbf{{Building and Room layer:}} Existing works, such as \cite{hughes2022hydra} and \cite{werby23hovsg}, have proposed various methods for room segmentation. We employed a prior-based method, partitioning objects in the offline map into distinct rooms. The resulting room layers are then combined to form the overall building layer.}

\textbf{{Carrier layer:}} We calculate the similarity between the text features $\textit{\textbf{T\_F}}_i$ of each object $\textit{\textbf{O}}_i$ and the SBERT-encoded text feature $\tilde{\textit{\textbf{T}}}$ for ``\textit{furniture for holding objects}''. The mathematical expression for this is as follows, \textcolor{black}{where $sim(\cdot,\cdot)$ denotes the cosine similarity function.}

\begin{equation}
\label{sbert_sim}
sim(\textit{\textbf{T\_F}}_i,\tilde{\textit{\textbf{T}}}) = \frac{\textit{\textbf{T\_F}}_i \cdot
 \tilde{\textit{\textbf{T}}}}{||\textit{\textbf{T\_F}}_i|| \cdot||\tilde{\textit{\textbf{T}}}||}
\end{equation}

Next, we select the set of objects $\tilde{\textit{\textbf{O}}} \subseteq \textit{\textbf{O}}$ with a similarity score exceeding a specified threshold $\sigma$, as shown below.

\begin{equation}
\label{sbert_thre}
 \tilde{\textit{\textbf{O}}} = \{ ({\textit{\textbf{O}}}_i,\textit{\textbf{cap}}_i,\textit{\textbf{T\_F}}_i) | sim(\textit{\textbf{T\_F}}_i,\tilde{\textit{\textbf{T}}}) > \sigma\}
\end{equation}

Next, we extract the three most frequent captions for each $\textit{\textbf{cap}}_i$ in $\tilde{\textit{\textbf{O}}}$, input them into a LLM (GPT-4o for test), and use a specific prompt to identify potential carrier-type objects, denoted as $\tilde{\textit{\textbf{O}}}_1 \subseteq \tilde{\textit{\textbf{O}}}$.

Finally, we select the finalized set of carrier-layer objects, denoted as $\bar{\textit{\textbf{O}}} 
\subseteq \tilde{\textit{\textbf{O}}}_1$, based on criteria such as the objects' geometric dimensions exceeding a certain size and their contact with the ground, as shown below.

\begin{equation}
\label{final_carrier}
 \bar{\textit{\textbf{O}}} \ = \ \textit{\textbf{f}} \ (\tilde{\textit{\textbf{O}}}_1)
\end{equation}

\textbf{{Carried layer and other objects:}} For any non-carrier-layer object $\textit{\textbf{O}}_i \in (\textit{\textbf{O}}-\bar{\textit{\textbf{O}}})$, \textcolor{black}{we determine whether $\textit{\textbf{O}}_i$ is carried by a carrier-layer object $\textit{\textbf{O}}_j \in \bar{\textit{\textbf{O}}}$ based on $\textit{\textbf{O}}_i$'s dimensions, the closest distance, and the spatial overlap relationship in the x-y-z directions between $\textit{\textbf{O}}_i$ and $\textit{\textbf{O}}_j$ (exceeding  a certain overlap rate).} $\textbf{\textit{h}}(\textit{\textbf{O}}_j, \textit{\textbf{O}}_i)$ is defined to encapsulate the consideration of the aforementioned factors, where 
$\textbf{\textit{h}}(\textit{\textbf{O}}_j, \textit{\textbf{O}}_i)=1$
if all the conditions are satisfied. For any $\textit{\textbf{O}}_j \in \bar{\textit{\textbf{O}}}$, we define the set of objects $\textbf{\textit{C}}(\textit{\textbf{O}}_j)$ carried by $\textit{\textbf{O}}_j$
 as follows:


 \begin{equation}
\label{smalls_on_carrier}
 \textbf{\textit{C}}(\textit{\textbf{O}}_j) = \{\textit{\textbf{O}}_i| \textbf{\textit{h}}(\textit{\textbf{O}}_j, \textit{\textbf{O}}_i)=1, \textit{\textbf{O}}_i \in  (\textit{\textbf{O}}-\bar{\textit{\textbf{O}}}) \}
\end{equation}

\textcolor{black}{All carried objects form the carried layer, represented by the object set $\mathcal{C}$ in Eq. \eqref{carried objects}. The carried-layer objects can be updated by adding or removing them based on the robot's latest environmental observations, as detailed in \ref{CRSG Adaptation}.}

\begin{equation}
\mathcal{C} = \bigcup_{\textit{\textbf{O}}_j \in \bar{\textit{\textbf{O}}}} C(\textit{\textbf{O}}_j)
\label{carried objects}
\end{equation}

\textcolor{black}{Except for the carrier-layer $\bar{\textit{\textbf{O}}}$ and carried-layer objects $\mathcal{C}$, the remaining objects are considered other objects, which can also be queried using text or CLIP features.} 

\subsection{Navigation Strategy for a Displaced Object}
Let the \textbf{input navigation command} for the target object be either a  $\textbf{\textit{text}}$, an $\textbf{\textit{image}}$, \textcolor{black}{or both}. \textcolor{black}{For solely an $\textbf{\textit{image}}$, it is passed through GPT-4o to obtain a textual description, denoted as $\textbf{\textit{text}}_1$, of the target object. Then, either $\textbf{\textit{text}}$ or $\textbf{\textit{text}}_1$ is encoded using the SBERT model. The resulting feature is compared with the SBERT features of each object in the CRSG $S\_G$ using cosine similarity, similar to Eq. \eqref{sbert_sim}. The object with the highest similarity score is selected as the target object, $\textbf{\textit{O}}_{target}$.}

We model the exploration of a displaced object as a fixed-policy Markov decision process (MDP) below.

\textbf{State Space $S$:} In the current step $t$, we define:

\textbf{1.} the robot's pose $L_t \in \mathcal{L}$, 

\textbf{2.} the set of unexplored carrier-layer objects $CR_t \in \mathcal{CR}$, 

\textbf{3.} the set of candidate target objects \textit{on the unexplored carrier-layer objects} $CT_t \in \mathcal{CT}$,

\textbf{4.} the flag of finding the target or not $F_t \in \{0,1\}$. ( $\mathcal{L}$, $\mathcal{CR}$ and $\mathcal{CT}$ denote the value set of $L_t$, $CR_t$ and $CT_t$ respectively.) 

The state variable $S_t$ is defined in Eq. \eqref{state_variable}.

 \begin{equation}
\label{state_variable}
 S_t = (L_t, CR_t, CT_t, F_t) \in S
\end{equation}

In the initial state $S_0 = (L_0, CR_0, CT_0, F_0)$, $L_0$ is the initial position of the robot, $CR_0 = \bar{\textit{\textbf{O}}}$, and $CT_0=\textbf{\textit{O}}_{target}$.

\textbf{Action Space $A$:}
\begin{equation}
\label{action space}
 A = \{Stop, Explore(cr), Goto(ct) \ | \ cr \in CR_t, ct \in CT_t  \}
\end{equation}

$Stop$ indicates that the task is completed or all carrier-layer objects have been explored. $Explore(cr)$ and $Goto(ct)$ represent exploring the carrier-layer object $cr \in CR_t$ and navigating to the location of $ct \in CT_t$, respectively.

The robot selects the next action $a_t \in A$ based on the current state $S_t$ according to a specific policy $\pi(\cdot)$ in Eq. \eqref{action_policy}.

\begin{equation}
\label{action_policy}
 a_t = \pi(S_t)
\end{equation}

\textbf{Policy $\pi(\cdot)$:}
Given current state $S_t = (L_t, CR_t, CT_t, F_t)$,   

\textbf{1.} if $F_t = 1$ or $CR_t=\emptyset$, then $a_t = Stop$. 

\textbf{2.} If $F_t = 0$ and $CT_t \ne \emptyset$,  we prioritize and select a candidate object to proceed with. Specifically, let $CT_t=\{ \textbf{\textit{O}}_{t1},..., \textbf{\textit{O}}_{ti} \}$. Some additional variables are stored: the SBERT similarities $SS_t=\{ \textbf{\textit{ss}}_{t1},..., \textbf{\textit{ss}}_{ti} \}$ between $CT_t$ and $\textbf{\textit{O}}_{target}$, the distances $D_t=\{ \textbf{\textit{d}}_{t1},..., \textbf{\textit{d}}_{ti} \}$ between $L_t$ and $CT_t$, and the average depth values $\tilde{D}_t=\{ \tilde{\textbf{\textit{d}}}_{t1},..., \tilde{\textbf{\textit{d}}}_{ti} \}$ when $CT_t$ are observed by the robot's camera. The priority rating of any $\textbf{\textit{O}}_{tj} \in CT_t$ corresponding to $\textit{\textbf{ss}}_{tj}$, $\textit{\textbf{d}}_{tj}$ and $\tilde{\textbf{\textit{d}}}_{tj}$, is evaluated as follows. 
\textcolor{black}{The parameters in \eqref{candidate_priority} and \eqref{f(d)} are set as $\omega_1=5$, $\omega_2=1$,  $\tilde{\textit{d}}_{1}=0.3$m, $\alpha=10$ and $\beta=0.1$ in the experiments.}


\begin{equation}
\label{candidate_priority}
\textcolor{black}{
 P\_R(\textbf{\textit{O}}_{tj}) = \textcolor{black}{\omega_r \cdot} \frac{\omega_1 \cdot \textit{\textbf{ss}}_{tj} \cdot f(\tilde{\textbf{\textit{d}}}_{tj})}{1 + \omega_2 \cdot \textbf{\textit{d}}_{tj}}  
 }
\end{equation}


\begin{equation}
\label{f(d)}
\textcolor{black}{
 f(\tilde{\textbf{\textit{d}}}_{tj}) 
 =
    \begin{cases}
    \ \ \text{exp}(\alpha(\tilde{\textbf{\textit{d}}}_{tj} - \tilde{\textit{d}}_{1}))   \ \ \ \ \ \ \tilde{\textbf{\textit{d}}}_{tj} < \tilde{\textit{d}}_{1}
    \\
   \ \ \text{exp}(-\beta(\tilde{\textbf{\textit{d}}}_{tj} - \tilde{\textit{d}}_{1})) \ \ \ \ \tilde{\textbf{\textit{d}}}_{tj} \ge \tilde{\textit{d}}_{1}
    \end{cases}}
\end{equation}
where  $\textit{\textbf{ss}}_{tj}$ is positively correlated with $P\_R(\textbf{\textit{O}}_{tj})$, as we assume that a larger $\textit{\textbf{ss}}_{tj}$ indicates a higher likelihood that the candidate is the target. \textcolor{black}{Moreover,
$f(\tilde{\textbf{\textit{d}}}_{tj})$ is used to model the confidence level of $\textit{\textbf{ss}}_{tj}$ based on $\tilde{\textbf{\textit{d}}}_{tj}$. Specifically, in Eq. \eqref{f(d)}, when $\tilde{\textbf{\textit{d}}}_{tj}$ is less than a small distance $\tilde{\textit{d}}_{1}$, we assume that the confidence of the front-end detection results drops sharply, as being too close to the detected object may cause observation distortion or incompleteness. Conversely, when $\tilde{\textbf{\textit{d}}}_{tj}$ exceeds this distance, we assume that the confidence decreases more gradually.} \textcolor{black}{Additionally, we generally assume that the moved target objects remain within the same room. Therefore, based on this assumption, $\omega_r=1$ if the two objects belong to the same room, and $\omega_r=0.8$ if they do not.} The robot will navigate to the location of the object with the maximum $P\_R$ and explore for $\textbf{\textit{O}}_{target}$.

\textbf{3.} \textcolor{black}{If $F_t = 0$, $CT_t = \emptyset$, and $CR_t \ne \emptyset$, the LLM selects one of the carrier objects $cr_{k} \in CR_t$ and the robot executes the action $a_t=Explore(cr_{k})$.} Specifically, the captions for each carrier object in $CR_t$ are extracted and provided as input to the LLM, along with the image or caption of the target object. Leveraging the LLM's commonsense understanding of carrier-carried relationships (e.g., ``\textit{a cup is unlikely to be placed on a toilet}''), the LLM identifies the carrier object where the target object is most likely to be found. \textcolor{black}{Additionally, the same-room principle is incorporated into the prompt given to the LLM.}

\textbf{State Transition Process:} 
 If $a_t =Explore(cr)$  (where $cr \in CR_t$) or  $a_t =Goto(ct)$  (where $ct \in CT_t$),  then  during the robot's movement, let $CR_{observed}$ represent the set of carrier objects observed within a small radius $r$ that have no candidate targets on them (based on the latest environmental observations), and let $CT_{new}$ represent the set of new target candidates found on unexplored carrier objects. Since some candidates in $CT_t$ may be carried by objects in $CR_{observed}$, $CT_t$ is updated to ${CT_t}^*$ after these candidates are removed. Specifically, the candidates in $CT_{new}$ are those for which the SBERT feature similarities with the target exceed a threshold $\sigma_1$. Additionally, the similarities between the target $\textbf{\textit{O}}_{target}$ and the objects carried in $CR_{observed}$ don't exceed $\sigma_1$.

\textbf{1.} if  $a_t=Explore(cr)$, $CR_{t+1}$ and $CT_{t+1}$ are updated as:
\begin{subequations}
\begin{align}
\label{CR_ex_t+1_}
 CR_{t+1} &= CR_t \ \backslash \ (\{cr\} \cup CR_{observed})  \\
 \label{CT_ex_t+1_}
 CT_{t+1} &= {CT_t}^*  \cup  CT_{new}
\end{align}
\end{subequations}

\textcolor{black}{\textbf{2.}} if \ $a_t=Goto(ct)$, $CR_{t+1}$ and $CT_{t+1}$ are updated as:
\begin{subequations}
\begin{align}
\label{goto_CR_t+1}
 CR_{t+1} &= CR_t \ \backslash \ (\{cr_1\} \cup CR_{observed}) , \ ct \in  \textbf{\textit{C}}(cr_1)
 \\
 \label{goto_CT_t+1_}
 CT_{t+1} &= {CT_t}^*  \cup  CT_{new} \ \backslash \ \{ct\}
\end{align}
\end{subequations}

\begin{figure}[t!]
\centering    
{
	\includegraphics[width=8cm]{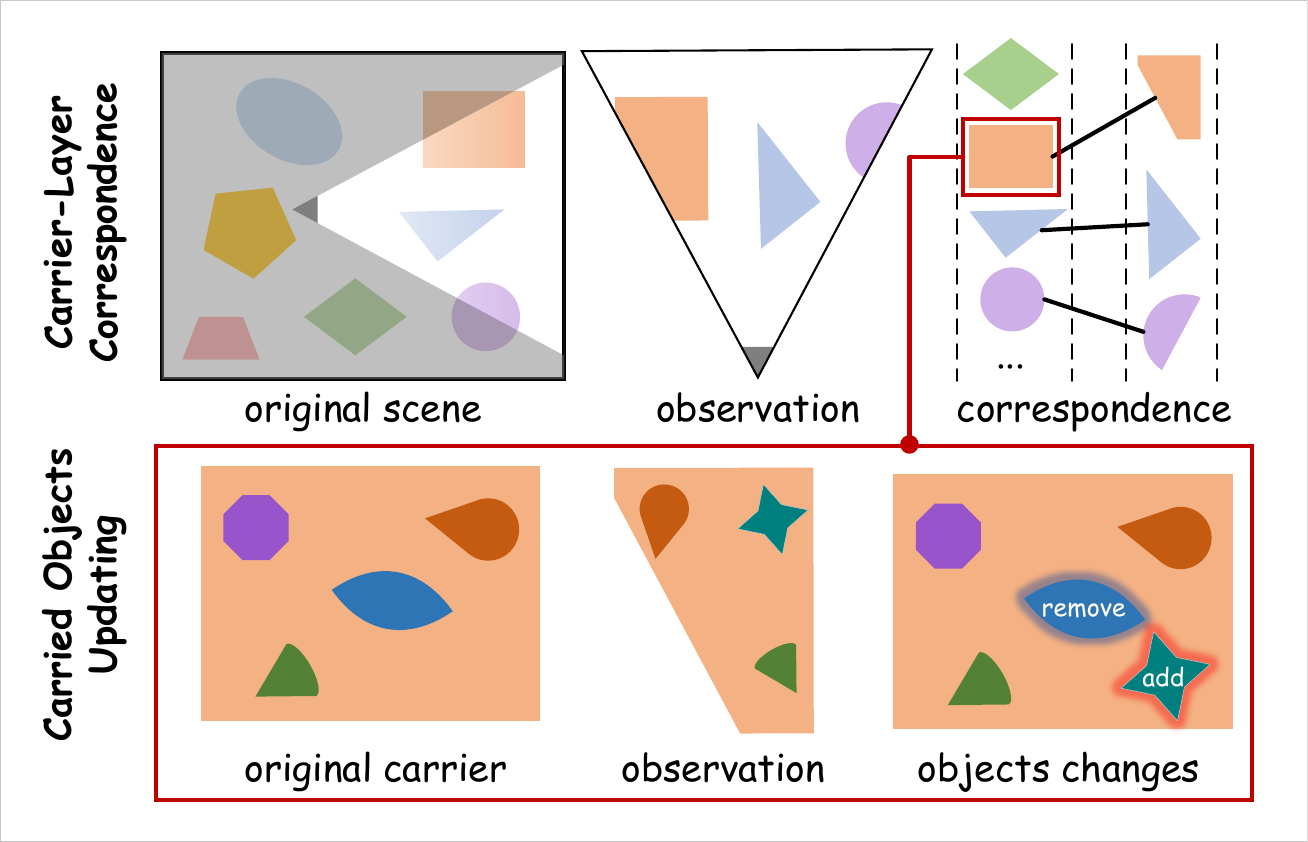} 

}
\caption{\textcolor{black}{A top-down schematic of CRSG Adaptation (\ref{CRSG Adaptation}).}} %
\label{updating} 
\end{figure} 

\textcolor{black}{\textbf{Target Determination Criteria.} In either case,  the similarities between $\textbf{\textit{O}}_{target}$ and objects carried by $cr$ or $cr_1$ are compared. First, we calculate the SBERT feature similarity $sim_{sbert}$ like Eq. \eqref{sbert_sim} between the target object and the carried objects. If the input includes an image, two more comparisons are made: \textbf{1.} A GPT-4o based image-to-image comparison is conducted, yielding  a probability $sim_{GPT}$
 that the objects are the same. \textbf{2.} We also compute the RGB histogram feature similarity $sim_{RGB}$ between the input $\textbf{\textit{image}}$ and the image of the carried object, denoted as $\textbf{\textit{image}}_c$, as follows.}

\begin{figure*}[t!]
\centering
{\includegraphics[width =16cm]{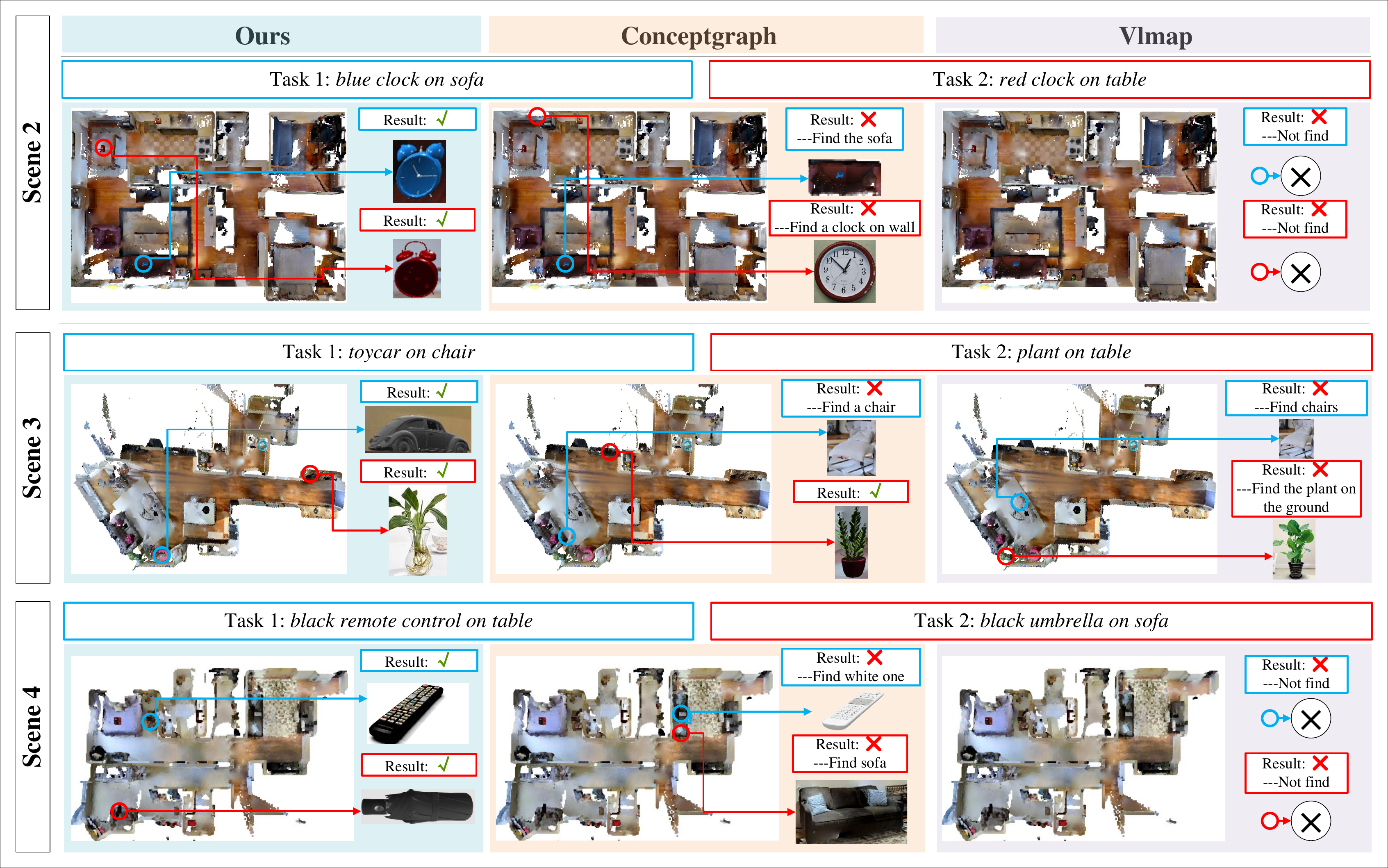}}
\caption{
Query Results for Some Carried Instances on the Offline Map.
}
\setlength{\belowcaptionskip}{-0.2cm}
\label{expe1}
\end{figure*}

\textcolor{black}{For each channel ($R$, $G$, $B$), the pixel values ranging from 0 to 255 are divided into $K$ intervals, and the number of pixels in each interval is counted. The histograms for each channel in an image are denoted as $\mathrm{hist}_R$, $\mathrm{hist}_G$ and $\mathrm{hist}_B$. The histogram for each channel is calculated below:}
\begin{equation}
\label{histogram}
\textcolor{black}{\mathrm{hist}_C[i] = \sum_{(x, y) \in image} \mathbb{I} \left( i \cdot \frac{256}{K} \leq C(x, y) < (i + 1) \cdot \frac{256}{K} \right)}
\end{equation}
\textcolor{black}{where $C \in \{R,G,B\}$, $i$ denotes the $i$-th interval and $\mathbb{I}(\cdot)$ is the indicator function, which equals 1 if the condition is true and 0 otherwise.}

\textcolor{black}{After merging the three histograms and normalizing, we obtain the final histograms of $\textbf{\textit{image}}$ and $\textbf{\textit{image}}_c$ denoted as $\mathrm{hist}_\textbf{\textit{image}}$ and $\mathrm{hist}_{\textbf{\textit{image}}_c}$. Finally, the calculation of $sim_{RGB}$ is based on the definition of $sim(\cdot,\cdot)$ as presented in Eq. \eqref{sbert_sim}.}

 \textcolor{black}{The determination of whether a carried object is the target object relies on a comprehensive evaluation of the values $sim_{sbert}$, $sim_{GPT}$ and $sim_{RGB}$. If the criteria are met, then $F_{t+1}=1$, and the task is marked as complete.} 

\subsection{\textcolor{black}{CRSG Adaptation}}
\label{CRSG Adaptation}

\textbf{Matching carrier-layer objects.} As the robot navigates, it periodically captures RGB and depth images from the environment. The RGB images are processed through CropFormer\cite{qi2022high}, Tokenize Anything\cite{pan2023tokenize}, CLIP\cite{radford2021learning} and SBERT\cite{reimers2019sentence} to obtain instance masks, captions, encoded CLIP features and SBERT features, respectively.  For newly observed objects $\textit{\textbf{O}}_{new}$, the robot compares them with a subset, $\bar{\textit{\textbf{O}}}_{sub}$, of the carrier-layer objects $\bar{\textit{\textbf{O}}}$ in $S_G$ to identify the observed carrier-layer objects, $\textit{\textbf{O}}_{match}^{cr}$.  The primary aspects of comparison between $\textit{\textbf{O}}_{new}$ and $\bar{\textit{\textbf{O}}}_{sub}$ include the object's size, the distance between their center positions, and the similarity scores based on SBERT features. \textcolor{black}{The subset $\bar{\textit{\textbf{O}}}_{sub}$ is selected from $\bar{\textit{\textbf{O}}}$ to improve the efficiency of CRSG updates, where each object in $\bar{\textit{\textbf{O}}}_{sub}$ satisfies the following criteria: }

\begin{equation}
    \label{subset of carried objects}
    \overrightarrow{(x_{yaw},y_{yaw})} \cdot \overrightarrow{(x_i-x_r,y_i-y_r)} \geq 0
\end{equation}
where $\overrightarrow{(x_{yaw},y_{yaw})}$, $(x_i,y_i)$, $(x_r,y_r)$ represent the robot's heading vector, the coordinates of any carrier-layer object, and the robot's current position, respectively.

\textbf{the Addition and Removal of Carried Objects.} For currently observed instances, $\textbf{\textit{h($\cdot,\cdot$)}}$ in Eq. \eqref{smalls_on_carrier} is used to determine whether they are being carried by $\textit{\textbf{O}}_{match}^{cr}$. Let the set of carried objects in the new observations be defined as $\textit{\textbf{O}}^{crd}$. The previously carried objects on $\textit{\textbf{O}}_{match}^{cr}$ are then compared with $\textit{\textbf{O}}^{crd}$. The comparison criteria include the object's size, the distance between their center positions, and the SBERT feature similarity score. After the comparison, the carried objects on $\textit{\textbf{O}}_{match}^{cr}$ are updated accordingly: they are either added, removed, or left unchanged. 
\textcolor{black}{Notably, certain carrier-layer objects are often only partially observed. Therefore, for each observed carrier-layer object, we calculate the distance between its point cloud and the carried objects recorded in the CRSG $S\_G$; if the distance exceeds a threshold, the carried object's state is not updated. Fig. \ref{updating} illustrates a schematic of the CRSG adaptation from a top-down view.}

\begin{table}[t!]
\centering
\caption{Success rate of object Query on the offline map.}
\label{tab1}
\resizebox{\linewidth}{!}{
\begin{tabular}{ccccccc}
\toprule
Method & scene\_1& scene\_2& scene\_3 & scene\_4 & scene\_5 &average\\ \midrule
 \textbf{Vlmap}\cite{huang2023visual}& 7/14& 8/19& 7/17 &  9/17 & 6/18&43.5\%\\
\textbf{Conceptgraph}\cite{gu2024conceptgraphs}& 9/14& 10/19&12/17 & 6/17& 8/18&52.9\%\\
\textbf{Ours}& \textbf{13/14}& \textbf{15/19}&\textbf{15/17} &\textbf{14/17} &\textbf{13/18} &\textbf{82.4\%}\\
\bottomrule
\end{tabular}
}\end{table}

\begin{table*}[t!]
\scriptsize
\centering
\caption{\textcolor{black}{SR and Tasks\_SR(\textit{i}) in different scenes for a series of long-sequence frequently used daily items navigation tasks, with the best in \colorbox{tableone}{pink} and the second best in \colorbox{tabletwo}{yellow}}.}
\label{exp-2}
\renewcommand\arraystretch{1.2}
\setlength{\tabcolsep}{1.7mm}
\textcolor{black}{
\begin{tabular}{c|c|ccccc|ccccc|ccccc|ccccc}
\toprule
\multicolumn{2}{c}\textbf{Scenes} & 
\multicolumn{5}{|c|}{\textbf{2}} & 
\multicolumn{5}{c|}{\textbf{3}} & 
\multicolumn{5}{c|}{\textbf{4}}& 
\multicolumn{5}{c}{\textbf{5}} \\ 
\midrule
\multicolumn{2}{c}\textbf{Methods} & 
\multicolumn{1}{|c}{\rotatebox{90}{\textbf{Vlfm}}} & \textbf{\rotatebox{90}{Ofm-0.4}} & \textbf{\rotatebox{90}{Ofm-0.55}} & \textbf{\rotatebox{90}{Ofm-0.7}}  & \textbf{\rotatebox{90}{Ours}}   & \textbf{\rotatebox{90}{Vlfm}} & \textbf{\rotatebox{90}{Ofm-0.4}} & \textbf{\rotatebox{90}{Ofm-0.55}} & \textbf{\rotatebox{90}{Ofm-0.7}}  & \textbf{\rotatebox{90}{Ours}} &  \textbf{\rotatebox{90}{Vlfm}} & \textbf{\rotatebox{90}{Ofm-0.4}} & \textbf{\rotatebox{90}{Ofm-0.55}} & \textbf{\rotatebox{90}{Ofm-0.7}}  & \textbf{\rotatebox{90}{Ours}}  & \textbf{\rotatebox{90}{Vlfm}} & \textbf{\rotatebox{90}{Ofm-0.4}} & \textbf{\rotatebox{90}{Ofm-0.55}} & \textbf{\rotatebox{90}{Ofm-0.7}}  & \textbf{\rotatebox{90}{Ours}}
\\ \midrule
\multirow{5}{*}{\rotatebox{90}{\textbf{SR (\%)}}} & \textbf{1} & \cellcolor{tabletwo}{40.0} & \cellcolor{tabletwo}{40.0} & 20.0 &  20.0 &  \cellcolor{tableone}{85.7} &  20.0 & 00.0 & 20.0 & \cellcolor{tabletwo}{60.0}  & \cellcolor{tableone}{80.0} &  00.0 &  00.0 & 00.0 & \cellcolor{tabletwo}{40.0}  & \cellcolor{tableone}{83.3} &  00.0 & \cellcolor{tabletwo}{20.0} & \cellcolor{tabletwo}{20.0} & \cellcolor{tableone}{40.0} & \cellcolor{tableone}{40.0}\\
& \textbf{2} & \cellcolor{tabletwo}{40.0} &  00.0 & \cellcolor{tabletwo}{40.0} &  20.0 & \cellcolor{tableone}{85.7} &  00.0 & 00.0 & \cellcolor{tableone}{100.0} & \cellcolor{tabletwo}{40.0}  &  \cellcolor{tabletwo}{40.0} & \cellcolor{tabletwo}{20.0} &  \cellcolor{tabletwo}{20.0} & 00.0 & 00.0  & \cellcolor{tableone}{83.3} &  00.0 & 40.0 & \cellcolor{tabletwo}{60.0} & \cellcolor{tableone}{80.0} & \cellcolor{tableone}{80.0}\\
& \textbf{3} & 20.0 &  20.0 & 20.0 & \cellcolor{tabletwo}{40.0}  & \cellcolor{tableone}{57.1} & \cellcolor{tabletwo}{40.0} & 00.0 & 00.0 & 20.0  &  \cellcolor{tableone}{100.0} & 00.0 &  \cellcolor{tabletwo}{40.0} & 20.0 & 00.0  & \cellcolor{tableone}{66.7}   &  00.0 & \cellcolor{tabletwo}{80.0} & 60.0 & 60.0 & \cellcolor{tableone}{100.0}\\
& \textbf{4} & 20.0 & 20.0 & \cellcolor{tabletwo}{40.0} & \cellcolor{tabletwo}{40.0}  &  \cellcolor{tableone}{85.7} &  40.0 &  \cellcolor{tabletwo}{60.0} & 40.0 & 40.0  &  \cellcolor{tableone}{80.0} & 00.0 &  \cellcolor{tabletwo}{20.0} & \cellcolor{tabletwo}{20.0} & 00.0  & \cellcolor{tableone}{83.3}  &  20.0 & 20.0 & \cellcolor{tableone}{60.0} & \cellcolor{tabletwo}{40.0} & \cellcolor{tabletwo}{40.0}\\
& \textbf{5} & 00.0 & \cellcolor{tabletwo}{33.3} & \cellcolor{tabletwo}{33.3} & \cellcolor{tabletwo}{33.3}  &  \cellcolor{tableone}{75.0} &  00.0 & \cellcolor{tabletwo}{66.7} & 33.3 & 33.3  &  \cellcolor{tableone}{100.0} & 00.0 &  \cellcolor{tabletwo}{25.0} & \cellcolor{tabletwo}{25.0} & 00.0  & \cellcolor{tableone}{75.0} &  00.0 & \cellcolor{tabletwo}{33.3} & \cellcolor{tabletwo}{33.3} & 00.0 & \cellcolor{tableone}{66.7}
\\ \midrule
\multirow{5}{*}{\rotatebox{90}{\textbf{Tasks\_SR(\textit{i}) (\%)}}} & \textbf{1} & \cellcolor{tabletwo}{40.0} & \cellcolor{tabletwo}{40.0} & 20.0 & 00.0  &  \cellcolor{tableone}{85.7} &  20.0 & 00.0 & 20.0 & \cellcolor{tabletwo}{60.0}  & \cellcolor{tableone}{80.0} &  00.0 &  00.0 & 00.0 & \cellcolor{tabletwo}{40.0}  & \cellcolor{tableone}{83.3} &  00.0 & \cellcolor{tabletwo}{20.0} & \cellcolor{tabletwo}{20.0} & \cellcolor{tableone}{40.0} & \cellcolor{tableone}{40.0}\\
& \textbf{2} & 00.0 &  00.0 & \cellcolor{tabletwo}{20.0} & 00.0  & \cellcolor{tableone}{71.4} &  00.0 & 00.0 & \cellcolor{tableone}{20.0} & \cellcolor{tableone}{20.0}  &  \cellcolor{tableone}{20.0} & 00.0 &  00.0 & 00.0 & 00.0  & \cellcolor{tableone}{83.3} &  00.0 & 00.0 & 00.0 & \cellcolor{tabletwo}{20.0} & \cellcolor{tableone}{40.0}\\
& \textbf{3} & 00.0 &  00.0 & 00.0 & 00.0  & \cellcolor{tableone}{57.1} &  00.0 & 00.0 & 00.0 & 00.0  &  \cellcolor{tableone}{20.0} & 00.0 &  00.0 & 00.0 & 00.0  & \cellcolor{tableone}{66.7}   &  00.0 & 00.0 & 00.0 & \cellcolor{tabletwo}{20.0} & \cellcolor{tableone}{40.0}\\
& \textbf{4} & 00.0 & 00.0 & 00.0 & 00.0  &  \cellcolor{tableone}{42.9} &  00.0 &  00.0 & 00.0 & 00.0  &  00.0 & 00.0 &  00.0 & 00.0 & 00.0  & \cellcolor{tableone}{66.7}  &  00.0 & 00.0 & 00.0 & \cellcolor{tableone}{20.0} & \cellcolor{tableone}{20.0}\\
& \textbf{5} & 00.0 & 00.0 & 00.0 & 00.0 &  \cellcolor{tableone}{50.0} &  00.0 & 00.0 & 00.0 & 00.0  &  00.0 & 00.0 &  00.0 & 00.0 & 00.0  & \cellcolor{tableone}{50.0} &  00.0 & 00.0 & 00.0 & 00.0 & 00.0\\
\bottomrule   
\end{tabular}
}
\end{table*}

\section{Experimental Results}\label{experiment}
By conducting extensive simulations and hardware experiments, we
 investigate the following key questions:

1. \textcolor{black}{Do the carried-by relationship and text description features improve the accuracy of instance queries? (Sec. \ref{Offline_Object_Query}})?

2. Does the dynamic update of the CRSG contribute to more efficient instance navigation (Sec. \ref{Long-sequence}, \ref{Ablation})? 

\textcolor{black}{3. Is our CRSG-based navigation strategy effective in  navigating to moved instances (Sec. \ref{Ablation})?}

\textbf{Metrics.}
We report Success Rate (SR) and Success weighted by inverse Path Length (SPL)\cite{anderson2018evaluation}. SPL measures the efficiency of a robot's path by comparing it to the shortest route from the starting point to the target object. If the robot fails to reach the target, the SPL is zero. Otherwise, SPL is the ratio of the shortest path length to the robot's actual path length, with higher values indicating better performance.

\subsection{Multi-type Query on the Offline Map}
\label{Offline_Object_Query}

\textbf{Baselines.}
\textcolor{black}{\textbf{Vlmap} \cite{huang2023visual} and \textbf{Conceptgraph} \cite{gu2024conceptgraphs} construct offline maps embedded with visual-language features for object queries and robot navigation, and are compared in terms of the accuracy of multi-type queries.}

\textcolor{black}{A total of 85 queries, covering different types of navigation instructions (semantic, instance, and demand-driven), were conducted across 5 scenes in Gibson \cite{xia2018gibson}, with each type accounting for 17.65\%, 49.41\%, and 32.94\% of the total, respectively.} The experimental results are presented in Tab. \ref{tab1}, where the query success rate of \textbf{Ours} averages 82.4\% and is the highest in all five scenes. Because in instance queries like ``\textit{a cup on the table}'', the CRSG of \textbf{Ours} records the carrying relationship between \textit{cup} and \textit{table}, allowing for precisely locating the instance. In contrast, \textbf{Vlmap}\cite{huang2023visual} and \textbf{Conceptgraph}\cite{gu2024conceptgraphs} may identify the \textit{table} instead of the \textit{cup}. Additionally, \textbf{Ours} additionally incorporates text features of caption descriptions for each instance in CRSG, enabling better differentiation between similar objects, such as \textit{black cup} and \textit{white cup}, and demonstrating superior performance in querying instances of specific colors. \textcolor{black}{Meanwhile, \textbf{Vlmap} projects CLIP features from 3D space to 2D grids, which can cause the loss of CLIP features for small objects. Additionally, its limited queryable semantic categories lead to inferior performance.} We illustrate partial query results in Fig. \ref{expe1}, and the results demonstrate that \textbf{Ours} outperforms in distinguishing between objects of the same category and in querying specific instances being carried.

\begin{figure}[t!]
\centering    
{
	\includegraphics[width=8.7cm]{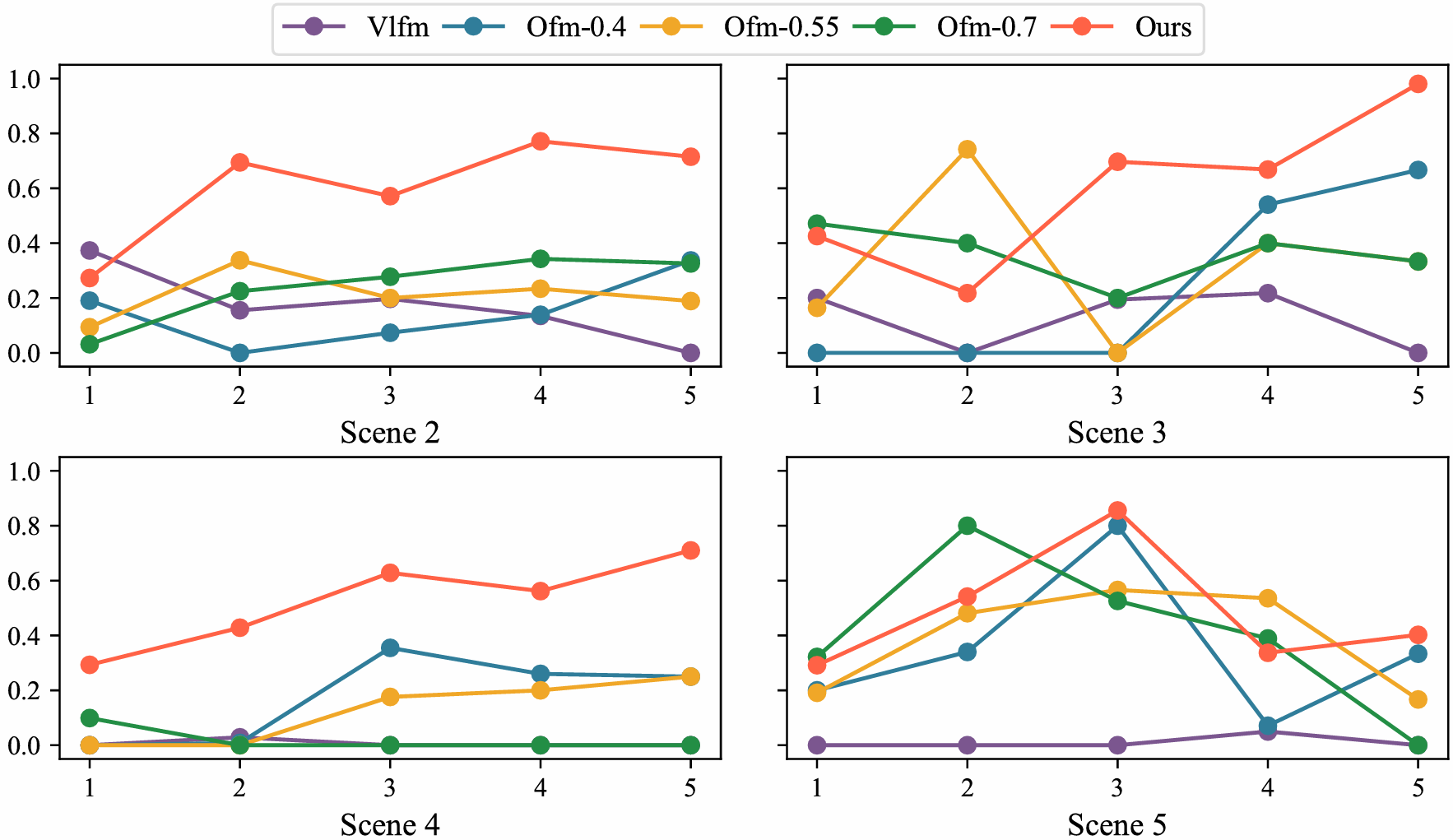} 

}
\caption{SPL of different methods for long-sequence object navigation. (\textit{horizontal-axis}: object number, \textit{vertical-axis}: SPL)} %
\label{spl} 
\end{figure}

\begin{figure*}[ht]
\centering
{\includegraphics[width =18cm]{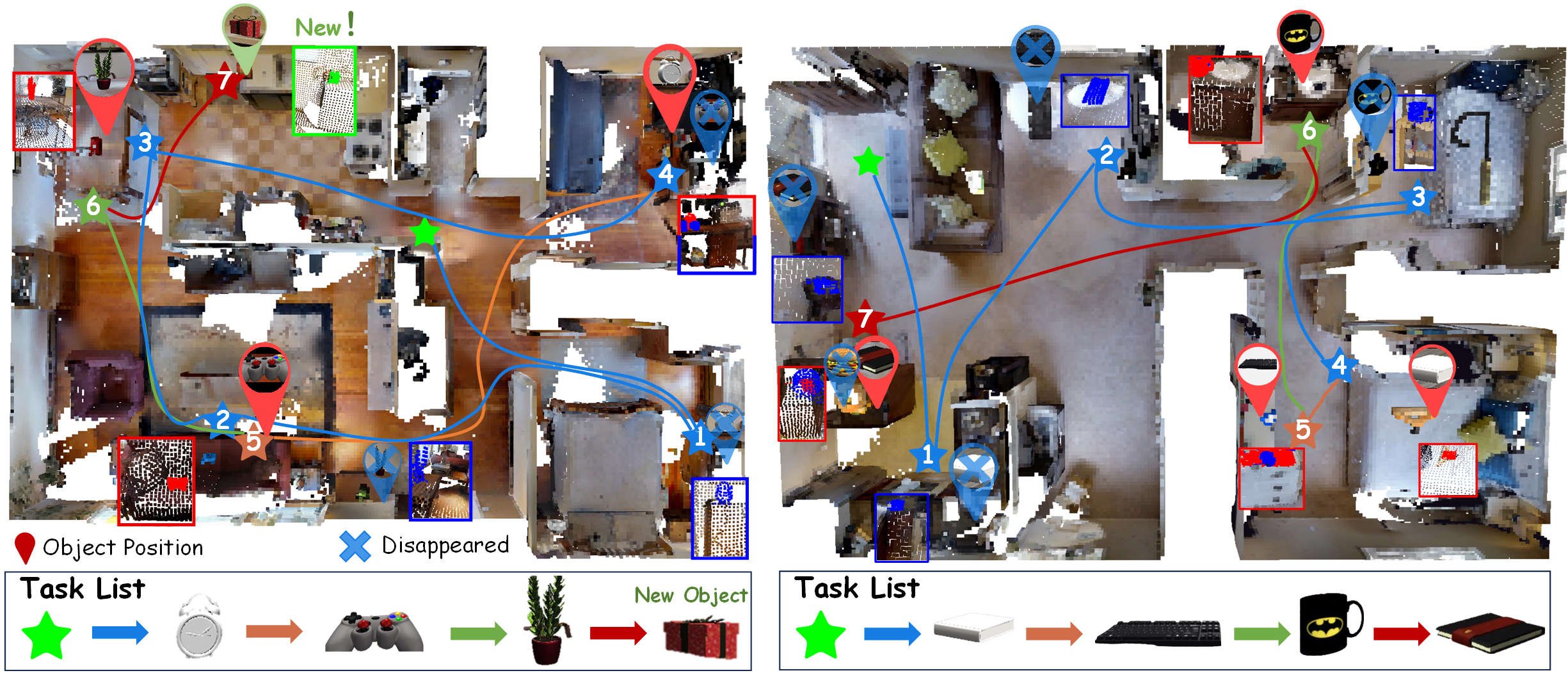}}
\caption{
The visualization showcases two long-sequence instance navigation results of \textbf{Ours}, with scene 2 on the left and scene 5 on the right. Updates to task-relevant objects in the CRSG map are highlighted in small frames: \textcolor{red}{red} objects within \textcolor{red}{red} borders indicate \textcolor{red}{appearances}, \textcolor{blue}{blue} objects within \textcolor{blue}{blue} borders denote \textcolor{blue}{disappearances}, and \textcolor{green}{green} objects within \textcolor{green}{green} borders represent \textcolor{green}{new additions} to the CRSG. During the initial exploration and navigation to the first object, the robot updated most of the CRSG area, enabling efficient navigation to subsequent objects by utilizing the known target locations.
}
\setlength{\belowcaptionskip}{-0.2cm}
\label{expe2-show}
\end{figure*}

\subsection{Long-sequence Navigation Task for Frequently Used Everyday
Instances}
\label{Long-sequence}
 
We conducted long-sequence navigation experiments (4–5 objects per sequence) in four everyday scenes from the Gibson dataset \cite{xia2018gibson} using the Habitat simulator \cite{savva2019habitat}. 
In each scene, we placed commonly used items (e.g., \textit{black cup}, \textit{blue clock}) along with distractor objects of the same category (e.g., \textit{black cup}, \textit{white cup}) to test accurate navigation to specific instances, and constructed an offline CRSG. Before the navigation experiments, the positions of these items were randomly altered to simulate the variability in their locations. The robot was instructed to navigate to these instances sequentially. 

\textbf{Baselines.} \textcolor{black}{The latest online exploratory open-vocabulary navigation methods, \textbf{Vlfm}\cite{yokoyama2024vlfm} and \textbf{Ofm}\cite{kuang2024openfmnav} (short for OpenFMNav) are selected as baselines. For \textbf{Ofm}\cite{kuang2024openfmnav}, it detects object semantics during navigation and records these semantics on a 2D grid map if the detection confidence exceeds a specific threshold. Due to the threshold's impact on navigation results, we conducted experiments with three thresholds: 0.4, 0.55 (official implementation), and 0.7, referred to as \textbf{Ofm-0.4}, \textbf{Ofm-0.55}, and \textbf{Ofm-0.7}, respectively. }

\textcolor{black}{The navigation results for each scene are shown in Tab. \ref{exp-2} (SR and Tasks\_SR(\textit{i})) and Fig. \ref{spl} (SPL), where Tasks\_SR(\textit{i}) represents the success rate of correctly navigating to all \textit{i} objects. Tab. \ref{exp-2} shows that \textbf{Ours} achieves the highest SR and Tasks\_SR(\textit{i}), thanks to its ability to identify instance-level targets. By supporting image input and integrating text features, RGB features, and VLM-based (GPT-4o) discriminations, \textbf{Ours} enables accurate navigation to the target instance. In contrast, both \textbf{Vlfm} and \textbf{Ofm} rely solely on text input. Specifically, \textbf{Vlfm} only supports navigation to general semantic categories, such as \textit{bottle}, rather than more specific instances like \textit{yellow bottle}, resulting in poor performance in correctly navigating to the target instance.} 

\textcolor{black}{In Fig. \ref{spl}, the SPL curves steadily increase with the number of objects, as the robot captures more changes in the scene and updates the CRSG throughout the long-sequence navigation task. As a result, the robot can often complete subsequent navigation tasks without further exploration. This highlights the importance of CRSG updates and showcases our scene maintenance capabilities. In contrast, \textbf{Vlfm} lacks scene maintenance, which prevents its SPL from improving as the long-sequence task progresses. Although \textbf{Ofm} has scene memory capabilities, its performance is limited by low success rates, leading to an SPL typically lower than \textbf{Ours}. Fig. \ref{expe2-show} illustrates two examples of long-sequence navigation using \textbf{Ours}, where the efficiency of target navigation improves significantly as the number of navigated objects increases.} 

\

\subsection{Ablation Study}
\label{Ablation}
\subsubsection{Different Criteria for Goal Determination}
\textcolor{black}{We conducted ablation experiments on all long-sequence navigation tasks in scene 4 to assess the necessity of using GPT-4o, text features, and RGB histogram features for determining whether the correct target has been navigated. The results, shown in Tab. \ref{goal-criteria-ablation}, highlight that \textbf{Ours} consistently achieves the highest success rate. The absence of any component—\textbf{w/o GPT-4o}, \textbf{w/o text}, or \textbf{w/o RGB}—results in lower performance, demonstrating the importance of all three components for accurate target navigation.}

\begin{table}[t!]
\centering
\caption{\textcolor{black}{Ablation Study 1: SR (\%) / Tasks\_SR(\textit{i}) (\%) Using Different Criteria for Goal Determination.}}
\label{goal-criteria-ablation}
\resizebox{0.9\linewidth}{!}{
\begin{tabular}{ccccc}

\toprule
Object & \textbf{w/o GPT-4o} & \textbf{w/o text} & \textbf{w/o RGB}  & \textbf{Ours}\\ \midrule
1 &66.7 / 66.7 & \textbf{83.3 / 83.3} & \textbf{83.3 / 83.3} & \textbf{83.3 / 83.3}\\ 
2 &66.7 / 66.7 & 50.0 / 50.0 & 50.0 / 50.0 & \textbf{83.3 / 83.3}\\
3 &\textbf{66.7} / 50.0 & \textbf{66.7} / 33.3 & \textbf{66.7} / 33.3 & \textbf{66.7 / 66.7}\\
4 &\textbf{83.3} / 50.0 & 50.0 / 33.3 & 50.0 / 33.3 & \textbf{83.3 / 66.7}\\ 
5 &50.0 / 00.0 & 25.0 / 25.0 & 25.0 / 25.0 & \textbf{75.0 / 50.0}\\
\bottomrule
\end{tabular}
}\end{table}

\begin{figure}[t]
\centering    
{
	\includegraphics[width=8 cm]{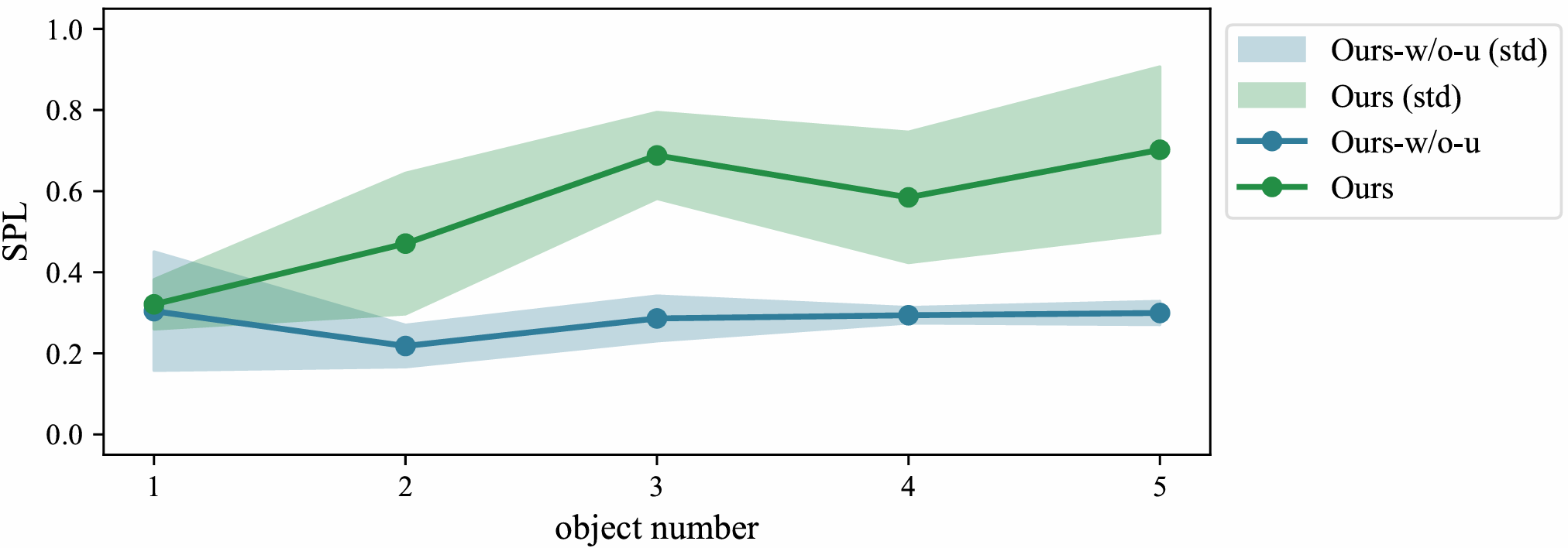} 

}
\caption{\textcolor{black}{Ablation Study 2: SPL of whether updating CRSG for long-sequence object navigation.}} %
\label{ablation2} 
\end{figure} 

\subsubsection{Impact of Updating CRSG} \textcolor{black}{We conducted an ablation study to evaluate the impact of CRSG updates on long-sequence navigation efficiency. A variant, \textbf{Ours-w/o-update}, relying on the initial CRSG for each object navigation task, was compared with \textbf{Ours} in terms of SPL across all scenes. Fig. \ref{ablation2} shows the average SPL, with shaded areas indicating the range of values. Without CRSG updates, the SPL of \textbf{Ours-w/o-update} does not increase as \textbf{Ours} does, highlighting the positive effect of CRSG updates on navigation efficiency.}

\subsubsection{Navigation Strategies}\textcolor{black}{We also conducted single daily instance navigation experiments in three scenes to evaluate the impact of various modules in our navigation strategy on navigation efficiency. \textbf{only-carriers\_Random} navigates to a randomly selected carrier object for exploration, without considering candidate target objects. \textbf{only-carriers\_LLM} builds on this by selecting the next carrier object based on LLM's recommendations. As shown in Tab. \ref{nav-strategy-ablation}, our method achieves the highest SPL, followed by \textbf{only-carriers\_LLM}, demonstrating that navigating to candidate target objects and selecting carrier objects using LLM’s commonsense knowledge improves navigation efficiency.}

\begin{table}[]
\centering
\caption{\textcolor{black}{Ablation Study 3: SPL by different navigation strategies.}}
\label{nav-strategy-ablation}

\resizebox{0.9\linewidth}{!}{
\begin{tabular}{cccc}
\toprule
Metric & \textbf{only-carriers\_Random} & \textbf{only-carriers\_LLM} & \textbf{Ours}\\ \midrule
SPL &0.205 & 0.309 & \textbf{0.342} \\
\bottomrule
\end{tabular}

}\end{table}

\begin{figure}[t!]
\centering    
{
	\includegraphics[width=8.5cm]{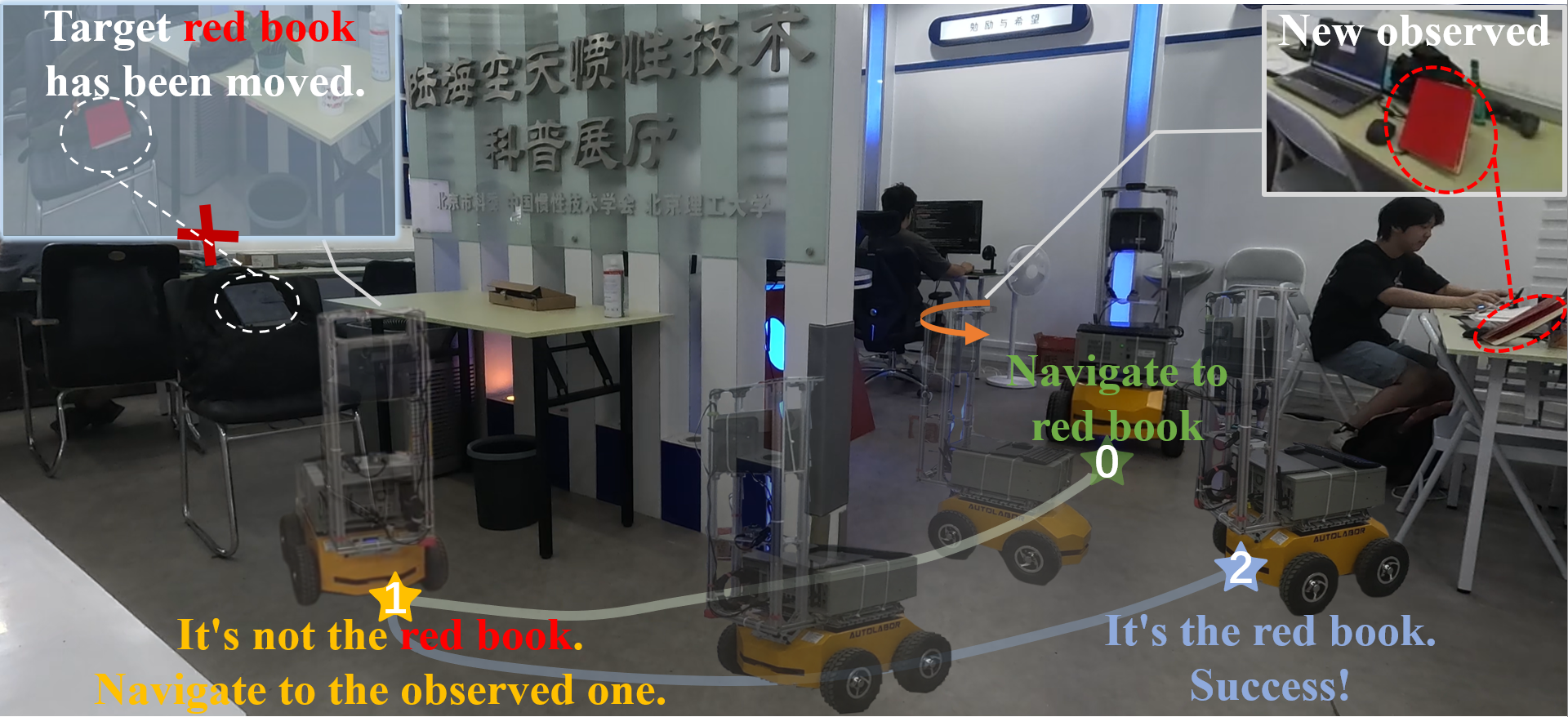} 

}
\caption{The robot queries the CRSG for the position of the \textbf{\textit{red book}} at the \textbf{\textit{chair}} and navigates there. It then discovers that the \textit{red book} is not in its original location and rules out the interference of the \textbf{\textit{grey book}}. Finally, it navigates to the \textbf{newly observed}  \textit{red book}.}%
\label{real} 
\vspace{-0.5cm}
\end{figure}

\subsection{Real-World Validation}
We validated our algorithm using an Autolabor robot in a real-world scene, equipped with an NVIDIA GeForce RTX 3090, Livox Mid 360 LiDAR for SLAM to obtain global poses, and an Azure Kinect DK for RGB-D capture. Path planning and obstacle avoidance were handled by the ROS move\_base package. The robot successfully navigated to a relocated \textit{red book}, while disregarding interference from a gray book (Fig. \ref{real}). 

\section{CONCLUSIONS} \label{CONCLUTIONS}
This paper has proposed an open-vocabulary navigation method for frequently used everyday items, leveraging a dynamic carrier-relationship scene graph (CRSG). Specifically, we first construct the CRSG to capture the dynamic relationships between carrier-level objects and the objects they carry. Next, a navigation strategy based on the CRSG is developed to navigate to frequently used items, modeling the object search process as a Markov Decision Process. At each navigation step, the CRSG is dynamically updated based on the robot's observations of the environment. The robot then decides to navigate toward candidate target objects or unexplored carrier objects, guided by visual-language feature similarities and commonsense knowledge from the LLM. Both simulations and hardware experiments demonstrate  that our method efficiently navigates to objects that are subject to positional changes, even in the presence of distractors from the same category.  \textcolor{black}{In the future, we plan to introduce an active exploration and mapping module to improve system scalability.}

\bibliographystyle{IEEEICRA}
\bibliography{ICRA}
\end{document}